\documentclass[11pt]{article}
\usepackage[preprint]{acl}

\usepackage{times}
\usepackage{latexsym}
\usepackage[T1]{fontenc}
\usepackage[utf8]{inputenc}
\usepackage{microtype}
\usepackage{inconsolata}
\usepackage{graphicx}
\usepackage{booktabs}
\usepackage{amsmath}
\usepackage{amssymb}
\usepackage{enumitem}
\usepackage{multirow}
\usepackage{xcolor}
\title{Displacement Is Not Direction: Evaluating Fidelity Metrics for Quantized LLM Deployment}

\author{
Milo\v{s} Nikoli\'{c}$^{1,2}$ \quad
Ali Hadi Zadeh$^{1,2}$ \quad
Enrique Torres Sanchez$^{1,2}$ \quad
Andreas Moshovos$^{1,2,3}$ \\
$^{1}$ByteShape \\
$^{2}$University of Toronto \\
$^{3}$Vector Institute for Artificial Intelligence \\
\texttt{contact@byteshape.com}
}

\begin{document}
\maketitle

\begin{abstract}
Fidelity metrics, such as per-token KL divergence (KLD) against a high-precision reference, are often used in practice as low-cost proxies for benchmark quality. We test this practice on a 28-quant cohort of Qwen3.6-35B-A3B and a 41-quant cohort of Devstral-Small-2-24B, evaluated across a suite of downstream benchmarks. We find that KLD is strongly correlated with benchmark score over the full cohort ($\rho=-0.72$ on Qwen and $\rho=-0.86$ on Devstral, both with $p<0.001$). However, this relationship collapses to non-significance in the near-baseline \emph{silent zone} ($\rho=+0.00$ on Qwen and $\rho=-0.24$, $p=0.36$, on Devstral). This collapse persists across 14 measurement variants, including different KLD aggregations, perplexity formulations, top-1 agreement, calibration corpora, and context lengths. At the per-prompt level, KLD has only weak failure-prediction power on code, with failed-vs-passed geometric-mean ratios in $[1.08,1.22]$ across five models on LiveCodeBench, and fails as a cross-model router, achieving only 42.3--49.4\% accuracy on disagreement prompts. We trace the collapse to a structural decomposition: KLD primarily measures the \emph{volume} of disagreement with the reference, with silent-zone composite $\rho=+0.94$ ($p<0.001$) on Qwen and $+0.55$ ($p=0.03$) on Devstral, while its relationship to the \emph{direction} of those disagreements is weak and task-conditional.

\end{abstract}

\section{Introduction}
\label{sec:intro}
As Large Language Models (LLMs) become more capable, their ability to solve real-world tasks increases the motivation for deploying them everywhere, from edge devices to datacenters. This capability is largely built on rapidly increasing parameter counts, ranging from the smallest several-billion-parameter models, all the way to trillion parameter ones. However, the ever larger models require more and more memory space and bandwidth, compute power and energy. 

One way to tackle the increasing size is to quantize and compress the model. This approach allows for squeezing out more throughput in data-centers, and is often the only way to deploy models on the edge. However, choosing a quantized model is no simple task. For Qwen3.6-35B-A3B \citep{qwen36_35b_a3b} alone, the community published hundreds of GGUF quants, made by different publishers, with different calibration recipes, and very different output qualities. Running a full benchmark suite on every candidate to judge quality is an expensive endeavour, so practitioners have resorted to using lower-cost metrics, in particular fidelity metrics like KL divergence (KLD) or Perplexity (PPL) against a BF16 reference.

This work examines the relationship between fidelity metrics, such as PPL and KLD, and downstream benchmark quality. At first glance, these metrics appear to work: across the full cohort, KLD and PPL correlate strongly with composite benchmark score, with Spearman $\rho(\text{KLD}, \text{composite quality}) = -0.72$ on Qwen3.6-35B-A3B and $-0.86$ on Devstral-2-Small-24B (both with $p<0.001$), and similar values across every fidelity-metric variant we tested.

However, this signal is not uniform across the quality range. It is driven primarily by clearly degraded quants, where KLD/PPL span a wide range and benchmark quality has already fallen substantially. When we restrict the analysis to near-baseline candidates, which are the models practitioners are most likely to choose from, the correlation collapses, with Spearman $\rho(\text{KLD}, \text{composite score}) = +0.00$ for Qwen and $-0.24$ ($p=0.36$) for Devstral. We refer to this near-baseline region as the \emph{silent zone}: KLD/PPL still measure distance from the reference model, but no longer provide useful ranking signal for downstream quality. In contrast, in the \emph{lossy zone}, KLD/PPL remain useful as coarse degradation detectors because the models have already suffered significant quality loss.

Inside the silent zone, fidelity metrics fail at three levels: they do not 1)~reliably rank models, 2)~separate correct from incorrect responses, or 3)~route a prompt to the better of two models when models disagree. We explain this with a volume--direction decomposition of score differences: KLD tracks how often the quantized model and reference differ in correctness, but task quality also depends on whether those differences are improvements or regressions.

In short, we make the following contributions:
\begin{enumerate}[itemsep=0pt,topsep=2pt]
    \item \textbf{Silent-zone vs Lossy-zone characterization: } We identify a near-baseline \textit{silent zone} where KLD/PPL lose ranking power, and a degraded lossy zone where they remain useful as coarse predictors (Table~\ref{tab:silent-zone}).
    \item \textbf{Decompose score into volume and direction: } We build on the work of \citep{dutta2024accuracy} by extending the observed flips into \textit{leapfrogs} (quantized model is correct where the reference is not) and \textit{drops} (the reverse), and derive a closed-form decomposition $\text{score} = \text{ref}_\text{score} + \text{vol} \cdot (2f - 1) / N$. KLD universally tracks volume; whether it also tracks direction is task-conditional (Table~\ref{tab:vol-dir}).
    \item \textbf{Per-prompt KLD has no useful model-selection signal: } At the per-prompt level on LiveCodeBench, KLD only weakly separates a model's own passes from its failures (fail-vs-pass geometric-mean ratios in $[1.08, 1.22]$ across five models) and fails to route between quants ($50.6-57.7\%$ error rate).
    \item \textbf{The collapse is metric-invariant: } The non-significant silent-zone correlation ($|\rho|\leq 0.25$, all $p>0.10$) holds across 14 variants of the per-token comparison metric, including PPL ratio, $\Delta$log-PPL, top-1 agreement, longer contexts, and on-task calibration (Table~\ref{tab:metric-invariant}).
\end{enumerate}

\section{Related Work}
\label{sec:related}

Quantization is widely used to reduce the memory, bandwidth, and compute cost of deploying large language models. To preserve quality at lower bit-widths, prior work has introduced practical post-training quantization methods such as GPTQ, AWQ, and SmoothQuant \citep{frantar2023optq, lin2023awq, xiao2023smoothquant}. Further compression can be achieved with mixed-precision quantization, where precision is assigned at finer granularity than the whole network, such as per tensor, layer, channel, or feature, e.g., \citep{zheng2024mixllm, bitpruning_iscas}. More recent studies evaluate the accuracy-performance trade-offs of quantized LLMs across model families, bit-widths, hardware targets, and downstream tasks \citep{lee2024comprehensive, jaiswal2024compressing, kurtic2025bf16}. These works show that quant quality cannot be summarized by model size alone, and that the practical value of a quantized model depends on the interaction between quality, throughput, hardware, and workload. Our work studies a narrower question: given multiple quantized variants of the same model, can inexpensive fidelity metrics such as KLD/PPL be used to rank their downstream task quality?

\subsection{Beyond PPL and KLD}

The limitations of PPL and related fidelity metrics for evaluating compressed models have been noted before. \citet{jaiswal2024compressing} show that pruned and quantized models can preserve perplexity while losing downstream capabilities, and introduce a broader benchmark suite for compressed models. \citet{deiseroth2024divergent} similarly argue that perplexity and aggregate accuracy do not fully capture generation degradation, and propose divergent-token metrics as a more generation-sensitive diagnostic for pruning and quantization. However, even though fidelity metrics are imperfect, their low computational cost keeps them attractive, especially in open-source model comparison. We build on this background, but focus specifically on model selection: whether fidelity metrics can distinguish among near-baseline quantized models, separate correct from incorrect responses, or route a prompt to the better of two quantized models.

\subsection{Flips}

\citet{dutta2024accuracy} present the closest prior art to our work. They introduce \textit{flips}: the percentage of multiple-choice answers that change correctness between a baseline model and a quantized model. They study the relationship between KLD, flips, and MMLU accuracy across quantized models, showing that accuracy alone can hide substantial behavior changes after quantization.

We extend this framework in five ways:
\begin{enumerate}[itemsep=0pt,topsep=2pt]
\item We extend the analysis beyond MMLU and multiple-choice accuracy to a broader set of tasks, including coding, tool calling, instruction following, and math reasoning.
\item We split flips into \emph{leapfrogs} (quantized model correct, reference wrong) and \emph{drops} (reference correct, quantized model wrong), and decompose score as $\text{ref}_\text{score} + \text{vol}\cdot(2f-1)/N$.
\item We show that KLD consistently predicts the volume of disagreement with the reference, but direction of those disagreements only conditionally on the task and quality regime.
\item We characterize the silent zone of near-baseline low-KLD quants, a regime their cohort does not isolate, where the apparent KLD-quality correlation evaporates.
\item We evaluate KLD at the per-prompt level and as a per-prompt routing signal, showing that on code benchmarks it fails both as a model-selection and response-correctness signal.
\end{enumerate}

Additionally, \citet{dutta2024accuracy} introduce top-margin analysis (the probability gap between the model's first and second choices). This is a per-prompt instability indicator. Our analysis focuses on whether fidelity to the reference predicts downstream task quality among quantized models.

\section{Methodology}
\label{sec:setup}

We evaluate two quantized-model cohorts with deliberately different architectures (Table~\ref{tab:setup}): 28 GGUF quantizations of Qwen3.6-35B-A3B~\citep{qwen36_35b_a3b}, a mixture-of-experts model, and 41 quantizations of the dense, code-specialized Devstral-Small-2-24B-Instruct-2512~\citep{devstral_small_2}. All fidelity metrics and leapfrog/drop counts are computed relative to the corresponding BF16 reference.

\begin{table}[t]
\centering
\caption{Two-cohort experimental setup. }
\label{tab:setup}

\setlength{\tabcolsep}{4pt}
\vspace{-5pt}
\resizebox{\linewidth}{!}
{
\begin{tabular}{lll}
\hline
\textbf{Item} & \textbf{Qwen cohort} & \textbf{Devstral cohort} \\\hline
Reference model     & Qwen3.6-35B-A3B & Devstral-Small-2-24B \\
Architecture        & MoE (35B/3B active) & dense (24B) \\
Specialization      & general / coding-capable & code-specialized \\
Cohort size         & 28 quants       & 41 quants \\
Quant sources       & 5               & 3 \\
bpw range           & 2.17--6.14      & 1.88--9.84 \\
Silent-zone $\tau$ & KLD $\leq 0.064$ & KLD $\leq 0.027$ \\
Silent / lossy split  & 17 / 11         & 16 / 25 \\
\hline
\end{tabular}
}
\vspace{-20pt}
\end{table}

The cohorts draw from multiple public quantization sources and span a broad range of compression, from near-reference high-bpw variants to aggressively compressed low-bpw ones. We use bits per weight (bpw) as a size proxy, not a quality metric: it sets the memory footprint, while benchmark scores and throughput determine practical quality.

\subsection{Downstream Quality Evaluation}

We evaluate downstream quality on benchmarks spanning coding (HumanEval~\citep{chen2021humaneval}, LiveCodeBench~v6~\citep{jain2024livecodebench}), instruction following (IFEval~\citep{zhou2023ifeval}), tool calling (BFCL~v3~\citep{patil2024bfcl}), math reasoning (GSM8K~\citep{cobbe2021gsm8k}), math and visual reasoning (GSM8K-v~\citep{gsm8kv}), and multitask knowledge (MMLU~\citep{hendrycks2021mmlu}); the Devstral cohort adds MATH-500~\citep{math500}. All benchmarks run in non-thinking mode, except GSM8K-v, which on the thinking-capable Qwen3.6 cohort we run in both modes.

We report per-benchmark scores and a composite (unweighted mean of normalized scores), oriented so higher is better and compared against the BF16 reference; per-prompt pass/fail outcomes feed the leapfrog/drop and routing analyses. Qwen MMLU is reported separately and excluded from the Qwen composite because of a prompt-format artifact, not a knowledge deficit (Appendix~\ref{sec:mmlu}). All evaluations use the EvalScope framework~\citep{evalscope_2024}, serving each GGUF model with \texttt{llama.cpp}~\citep{llama_cpp} (build b8855 for Qwen3.6, b7744 for Devstral).

\subsection{Fidelity Metrics}

Unless otherwise stated, KLD denotes $\mathrm{KL}(p_{\mathrm{BF16}} \,\|\, p_{\mathrm{quant}})$ over next-token distributions. Response-only measurements score only response tokens (excluding the prompt); two diagnostics use this form with top-$k$ truncation, defined in Appendix~\ref{sec:kld-topk}: the validation-set variants in Table~\ref{tab:metric-invariant} (a held-out instruction mixture, Appendix~\ref{sec:dataset}) and the per-prompt analyses (benchmark prompt-response pairs).

We compute PPL and KLD with \texttt{llama-perplexity} from the same builds. The default and headline metric is WikiText~\citep{merity2017wikitext} at 512-token context, which defines the main silent-zone split; we also test 8192-token context and the response-only variants above to probe sensitivity to context length and corpus. All other variants are diagnostic only.

We report results for the full cohort and a cohort-specific near-baseline subset, the \emph{silent zone}.

\section{The Silent Zone}
\label{sec:silent}

Our first finding is that every fidelity metric we tested exhibits a \emph{silent zone}: a near-baseline region where the metric no longer ranks downstream benchmark quality. Over the full cohort, KLD/PPL often correlate strongly with benchmark score. However, this signal is driven mainly by aggressively compressed models whose quality has already degraded.

For each cohort, we report a descriptive silent-zone threshold $\tau$ that marks the near-baseline cluster of quants: a KLD cutoff below which the within-zone Spearman correlation between KLD and composite quality is statistically non-significant ($p>0.10$), i.e., a regime with no detectable monotone KLD--quality signal. Models below $\tau$ form the silent zone; models above $\tau$ form the lossy zone. This threshold is descriptive, cohort-specific, and metric-specific. For default WikiText KLD at 512-token context, $\tau=0.064$ for Qwen and $\tau=0.027$ for Devstral. Because $\tau$ is not intended as a general decision rule, Appendix~\ref{sec:noncircular} repeats the analysis with a KLD-threshold-free quality definition, where near-baseline quants are those that retain at least $99\%$ of the BF16 composite score.

\begin{figure*}[t]
  \centering
  \vspace{-15pt}
  \begin{minipage}[t]{0.49\linewidth}
    \centering
    \includegraphics[width=\linewidth]{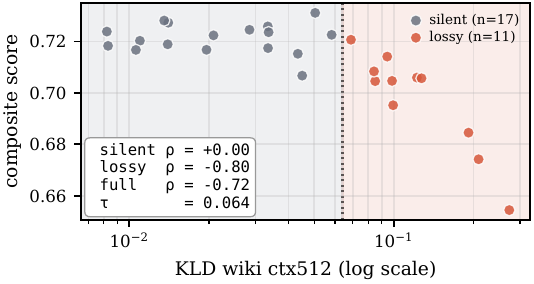}
\vspace{-25pt}
    \caption*{(a) Qwen3.6-35B-A3B}
  \end{minipage}\hfill
  \begin{minipage}[t]{0.49\linewidth}
    \centering
    \includegraphics[width=\linewidth]{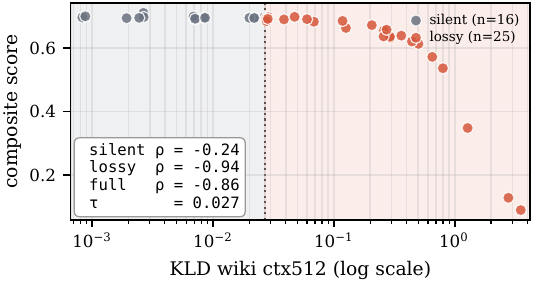}
\vspace{-25pt}
    \caption*{(b) Devstral-Small-2-24B}
  \end{minipage}
\vspace{-5pt}
  \caption{The silent-zone collapse on both cohorts, with Spearman correlation coefficients $\rho(\text{KLD}, \text{composite score})$ for the full cohort and the silent zone.}
  \label{fig:silent-zone}
\vspace{-15pt}
\end{figure*}

\subsection{The silent-zone collapse}
\label{sec:silent-collapse}

Figure~\ref{fig:silent-zone} shows the collapse on both cohorts. Over the full cohort, KLD is strongly negatively correlated with composite quality: $\rho=-0.72^*$ on Qwen and $\rho=-0.86^*$ on Devstral. Inside the silent zone, the correlation falls to $+0.00$ on Qwen (17 models at KLD $\leq 0.064$) and a non-significant $-0.24$ ($p=0.36$) on Devstral (16 models at KLD $\leq 0.027$). Thus, the full-cohort relationship does not imply that KLD can rank the near-baseline models a practitioner would normally compare. It mostly reflects the separation between near-baseline models and substantially degraded ones.

Two cohorts with different architectures (MoE vs.\ dense), training profiles (general vs.\ code-specialized), sizes (35B/3B-active vs.\ 24B), and publishers show the same qualitative collapse: the negative full-cohort signal commonly measured is driven primarily by degraded low-bit quants that show noticeable degradation from baseline quality. At or near baseline quality, benchmark score and KLD are essentially uncorrelated.

By contrast, the lossy-zone side shows a strong correlation on the other side: $\rho$ is $-0.80$ ($p=0.003$) for Qwen and $-0.94$ ($p<0.001$) for Devstral. This correlation is far stronger than the full-cohort numbers in either case.
What we usually \emph{see} as the full-cohort KLD-vs-quality correlation is the lossy zone leaking into the cohort average through the inclusion of 2--3-bit models that have noticeably degraded quality.

\begin{table}[t]
  \centering
  \small
  \setlength{\tabcolsep}{5pt}
  \caption{Spearman $\rho(\text{KLD}, \text{benchmark score})$ by zone, on both cohorts. $^*\, p < 0.05$, $^\dagger\, p < 0.10$.}
  \label{tab:silent-zone}
  \vspace{-5pt}
\resizebox{\linewidth}{!}
{
  \begin{tabular}{llll|lll}
    \toprule
    & \multicolumn{3}{c|}{\textbf{Qwen3.6-35B-A3B}} & \multicolumn{3}{c}{\textbf{Devstral-Small-2-24B}} \\
    \cmidrule(lr){2-4} \cmidrule(lr){5-7}
    \textbf{Benchmark} & \multicolumn{1}{c}{\textbf{silent
    }} & \multicolumn{1}{c}{\textbf{lossy}} & \multicolumn{1}{c|}{\textbf{full}} & \multicolumn{1}{c}{\textbf{silent}} & \multicolumn{1}{c}{\textbf{lossy}} & \multicolumn{1}{c}{\textbf{full}}\\
    \midrule
    HumanEval          & $+0.20$         & $-0.91^*$         & $-0.19$   & $-0.14$    & $-0.81^*$  & $-0.63^*$ \\
    LiveCodeBench      & $+0.08$         & $-0.87^*$         & $-0.65^*$ & $-0.20$    & $-0.96^*$  & $-0.92^*$ \\
    IFEval             & $+0.18$         & $-0.07$           & $-0.26$   & $-0.10$    & $-0.87^*$  & $-0.73^*$ \\
    BFCL\_v3           & $-0.24$         & $-0.25$           & $-0.71^*$ & $-0.12$    & $-0.90^*$  & $-0.75^*$ \\
    GSM8K  & $+0.11$         & $-0.61^*$         & $-0.43^*$ & $-0.05$    & $-0.96^*$  & $-0.89^*$ \\
    GSM8K-v            & $-0.43^\dagger$ & $-0.57^\dagger$   & $-0.83^*$ & $-0.14$    & $-0.88^*$  & $-0.74^*$ \\
    GSM8K-v think   & $-0.51^*$       & $-0.84^*$         & $-0.83^*$ & \multicolumn{1}{c}{---}        & \multicolumn{1}{c}{---}        & \multicolumn{1}{c}{---} \\
    MATH-500           & \multicolumn{1}{c}{---}             & \multicolumn{1}{c}{---}               & \multicolumn{1}{c|}{---}       & $+0.30$    & $-0.95^*$  & $-0.80^*$ \\
    MMLU               & $-0.54^*$       & $+0.50$           & $+0.39^*$ & $-0.22$    & $-0.98^*$  & $-0.89^*$ \\
    \midrule
    \textbf{Composite} & $+0.00$ & $-0.80^*$ & $-0.72^*$ & $-0.24$ & $-0.94^*$ & $-0.86^*$ \\
    \bottomrule
  \end{tabular}
  }
  
\vspace{-5pt}
\end{table}

Table~\ref{tab:silent-zone} breaks the pattern down per benchmark on both cohorts and shows that the collapse is across the board and not driven by any single benchmark.
On every benchmark, the lossy zone alone gives a strong negative $\rho$ (the degraded models are reliably worse on quality and reliably higher on KLD). On the other hand, in the silent zone the sign collapses to non-significant or wrong-signed on most benchmarks across both cohorts. The only benchmark exception with detectable silent-zone signal at $p<0.05$ is Qwen GSM8K-v thinking.

\subsection{Collapse is metric-invariant}
\label{sec:metric-invariant}

\begin{table}[t]
  \centering
  \footnotesize
  \setlength{\tabcolsep}{2pt}
  \caption{The silent-zone collapse holds across 14 measurement variants, on both cohorts.}
  \label{tab:metric-invariant}
  \vspace{-5pt}
  \resizebox{\linewidth}{!}
{
  \begin{tabular}{lrr|rr}
    \toprule
    & \multicolumn{2}{c}{\textbf{Qwen3.6-35B-A3B}} & \multicolumn{2}{c}{\textbf{Devstral-Small-2-24B}} \\
    \cmidrule(lr){2-3} \cmidrule(lr){4-5}
    \textbf{Metric variant} & \textbf{full} & \textbf{silent} & \textbf{full} & \textbf{silent} \\
    \midrule
    KLD mean wiki c512 & $-0.72^*$ & $+0.00$ & $-0.86^*$ & $-0.24$ \\
    KLD median wiki c512                  & $-0.69^*$  & $+0.03$ & $-0.86^*$  & $-0.24$ \\
    KLD p90 wiki c512                     & $-0.71^*$  & $+0.01$ & $-0.86^*$  & $-0.24$ \\
    KLD p99 (tail) wiki c512              & $-0.75^*$  & $-0.11$ & $-0.86^*$  & $-0.21$ \\
    mean $\Delta\log$-PPL wiki c512       & $-0.68^*$  & $+0.09$ & $-0.87^*$  & $-0.25$ \\
    PPL ratio (q/base) wiki c512          & $-0.68^*$  & $+0.09$ & $-0.87^*$  & $-0.25$ \\
    PPL diff wiki c512                    & $-0.68^*$  & $+0.09$ & $-0.87^*$  & $-0.25$ \\
    top-1 agreement \% wiki c512          & $+0.70^*$  & $-0.04$ & $+0.86^*$  & $+0.25$ \\
    KLD wiki c8k (long context)           & $-0.72^*$  & $+0.00$ & $-0.86^*$  & $-0.23$ \\
    PPL ratio wiki c8k                    & $-0.69^*$  & $+0.09$ & $-0.86^*$  & $-0.18$ \\
    top-1 agreement \% wiki c8k           & $+0.69^*$  & $-0.03$ & $+0.86^*$  & $+0.24$ \\
        KLD val-wiki c512                     & $-0.71^*$  & $+0.03$ & ---        & ---     \\
    KLD val-response top-$k$=60           & $-0.73^*$  & $-0.03$ & $-0.86^*$  & $-0.21$ \\
    KLD val-response top-$k$=20           & $-0.73^*$  & $-0.03$ & $-0.87^*$  & $-0.21$ \\
    \bottomrule
  \end{tabular}
  }
  
\vspace{-20pt}
\end{table}

A natural pushback is that the collapse may be specific to our KLD formulation. We retest the correlation across 14 variants spanning KLD aggregation, PPL formulation, top-1 agreement, context length, calibration corpus, top-$k$ truncation, and response-only measurement. Table~\ref{tab:metric-invariant} shows that all variants preserve the full-cohort signal but collapse in the silent zone, with $|\rho| \leq 0.25$ ($p > 0.10$) on both cohorts. Even response-only KLD on a task-aligned validation set shows the same pattern, suggesting that the collapse is not an artifact of WikiText, context length, or aggregation choice.

\subsection{Per-prompt KLD analysis}
\label{sec:per-prompt-null}

To further strengthen our analysis and give KLD the best opportunity, we remove all imperfect or unrepresentative dataset excuses and do a \emph{per-prompt} KLD experiment. Instead of calculating KLD as a dataset-aggregate scalar, we feed the model exact benchmark prompts, calculate KLD on the responses and compare to the benchmark pass/fail outcomes. If KLD had any predictive power in the \textit{silent-zone}, we should be able to identify which prompts a given quant will fail.

\begin{figure}[t]
  \centering
  \vspace{0pt}
  \includegraphics[width=\columnwidth]{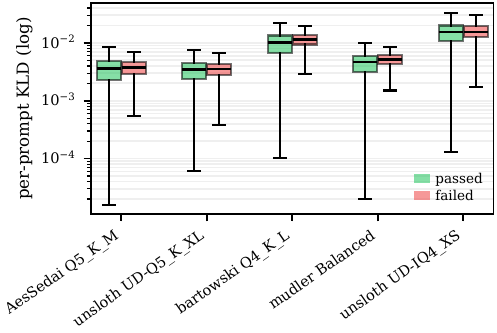}
\vspace{-25pt}
  \caption{Per-prompt KLD distribution on LiveCodeBench, split by each model's own pass (green) / fail (red) outcomes, for five silent-zone Qwen quants.}
  \label{fig:per-prompt}
\vspace{-20pt}
\end{figure}

We compute per-prompt KLD on LCB on the Qwen cohort for five silent-zone quants and split each model's prompts into the ones it passed and the ones it failed. We plot the KLD distribution for this mini cohort in Figure~\ref{fig:per-prompt}. There is a small failure-prediction signal, with the geometric-mean per-prompt KLD ranging between 1.08 and 1.22 ($13\%$ average increase). However, this ratio is far below what is needed for routing or screening.

A stronger test is whether per-prompt KLD can choose between two quantized models on the same prompt. We pair model B (ByteShape IQ4\_XS-3.93bpw, KLD $0.069$) against each of five clean baselines on LiveCodeBench and ask: on prompts where exactly one of the two passes, does routing to the lower per-prompt KLD model identify the passer more than half the time?

Table~\ref{tab:router} shows that the answer is no.
Lower-KLD-wins accuracy is in $[42.3\%, 49.4\%]$ across the five pairings, never reaching coin toss chance.
The signed correlation $\rho(\text{KLD}, \text{correctness})$ is positive (wrong-direction) in four and essentially zero ($\rho=-0.01$) in the fifth. On the BF16-fail slice (the prompts where a quant could leapfrog the reference), accuracy drops further to $40.8\%$.

\begin{table}[t]
  \centering
  \small
  \vspace{-15pt}
  \caption{Cross-model picker on LiveCodeBench disagreement prompts (exactly one of the two passes).}
  \vspace{-5pt}
  \label{tab:router}
  \resizebox{\linewidth}{!}
{
  \begin{tabular}{lccc}
    \toprule
    \textbf{Pairing (vs.\ B)} & \textbf{disagrees} & \textbf{wins} & \textbf{$\rho$ ($p$)} \\
    \midrule
    AesSedai-Q5\_K\_M    & 190 & $47.4\%$ & $+0.033\,(0.53)$ \\
    unsloth-UD-Q5\_K\_XL & 177 & $44.1\%$ & $+0.109\,(0.040)$ \\
    mudler-APEX-Balanced & 170 & $42.9\%$ & $+0.126\,(0.020)$ \\
    bartowski-Q4\_K\_L   & 163 & $42.3\%$ & $+0.128\,(0.020)$ \\
    unsloth-UD-IQ4\_XS   & 180 & $49.4\%$ & $-0.009\,(0.86)$ \\
    \midrule
    BF16-fails (AesSedai-Q5\_K\_M vs.\ B) & 71 & $40.8\%$ & $+0.159\,(0.059)$ \\
    \bottomrule
  \end{tabular}
  }
  \vspace{-15pt}
\end{table}

Per-prompt KLD contains a weak failure signal, but not a useful selection signal. It is not strong enough to reliably identify a model's failures, nor can it route between models better than chance.

\section{Disagreement Volume and Direction}
\label{sec:volume-direction}

What does KLD actually measure, if not quality?
We propose a decomposition that explains it.

For a benchmark of $N$ prompts where the reference (BF16) passes $N_p$ and a candidate quant passes $N_q$, define: (1) \textbf{leapfrogs} $\ell$: prompts where quant passes and reference fails; (2) \textbf{drops} $d$: prompts where quant fails and reference passes; (3) \textbf{volume} $\text{vol} = \ell + d$: total disagreement count; and (4) \textbf{direction} $f = \ell / \text{vol}$: fraction of disagreements that are improvements. By construction:
\begin{equation}
\vspace{-5pt}
  \text{score}_q \;=\; \text{score}_\text{ref} + \frac{\text{vol} \cdot (2f - 1)}{N}
  \label{eq:decomp}
  \vspace{-5pt}
\end{equation}
where $\text{score}_q$ and $\text{score}_{ref}$ are benchmark scores of the quant and BF16. Score is determined by \emph{how many} prompts the quant flips (volume) and \emph{in which direction} ($2f - 1$). That is: the volume controls magnitude, and the direction controls the sign. The identity is exact for benchmarks whose official score is one binary outcome per prompt (HumanEval, LCB, IFEval, GSM8K, GSM8K-v, MATH-500), and approximate for BFCL\_v3 and MMLU, whose official scores macro-average over sub-categories with a weighting different from the per-prompt count we use for leap/drop accounting.

\vspace{-0.5em}
\paragraph{KLD measures displacement, not direction: }
\begin{table}[t]
  \centering
  \footnotesize
  \vspace{-20pt}
  \caption{$\rho(\text{KLD}, \cdot)$ for volume and direction $f$ per benchmark and zone, on both cohorts.}
  \label{tab:vol-dir}
  \vspace{-5pt}
  \setlength{\tabcolsep}{2pt}
  \resizebox{\linewidth}{!}
{
  \begin{tabular}{lllll|llll}
    \toprule
    & \multicolumn{4}{c|}{\textbf{Qwen3.6-35B-A3B}} & \multicolumn{4}{c}{\textbf{Devstral-Small-2-24B}} \\
    \cmidrule(lr){2-5} \cmidrule(lr){6-9}
    & \multicolumn{2}{c}{silent} & \multicolumn{2}{c|}{full} & \multicolumn{2}{c}{silent} & \multicolumn{2}{c}{full} \\
    \cmidrule(lr){2-3} \cmidrule(lr){4-5} \cmidrule(lr){6-7} \cmidrule(lr){8-9}
    \textbf{Benchmark} & \multicolumn{1}{c}{vol} & \multicolumn{1}{c}{$f$} & \multicolumn{1}{c}{vol} & \multicolumn{1}{c|}{$f$} & \multicolumn{1}{c}{vol} & \multicolumn{1}{c}{$f$} & \multicolumn{1}{c}{vol} & \multicolumn{1}{c}{$f$} \\
    \midrule
    HumanEval         & $+0.51^*$       & $+0.02$            & $+0.76^*$ & $-0.31$           & $+0.73^*$ & $+0.35$            & $+0.91^*$ & $-0.17$ \\
    LiveCodeBench     & $+0.50^*$       & $+0.22$            & $+0.82^*$ & $-0.59^*$         & \multicolumn{1}{c}{---}      & \multicolumn{1}{c}{---}                & \multicolumn{1}{c}{---}       & \multicolumn{1}{c}{---} \\
    IFEval            & $+0.42^\dagger$ & $+0.17$            & $+0.67^*$ & $-0.18$           & $+0.20$   & $-0.15$            & $+0.81^*$ & $-0.71^*$ \\
    BFCL\_v3          & $+0.91^*$       & $-0.14$            & $+0.96^*$ & $+0.44^*$         & $+0.51^*$ & $+0.17$            & $+0.84^*$ & $-0.57^*$ \\
    GSM8K             & $+0.38$         & $+0.04$            & $+0.59^*$ & $-0.48^*$         & $+0.58^*$ & $-0.04$            & $+0.92^*$ & $-0.82^*$ \\
    GSM8K-v           & $+0.62^*$       & $-0.35$            & $+0.88^*$ & $-0.81^*$         & $+0.55^*$ & $-0.31$            & $+0.90^*$ & $-0.77^*$ \\
    GSM8K-v think     & $+0.66^*$       & $-0.52^*$ & $+0.84^*$ & $-0.83^*$         & \multicolumn{1}{c}{---}      & \multicolumn{1}{c}{---}                & \multicolumn{1}{c}{---}       & \multicolumn{1}{c}{---} \\
    MATH-500          & \multicolumn{1}{c}{---}      & \multicolumn{1}{c}{---}                & \multicolumn{1}{c}{---}       & \multicolumn{1}{c|}{---} & $+0.19$   & $+0.38$            & $+0.78^*$ & $-0.80^*$ \\
    MMLU              & $+0.66^*$       & $-0.27$            & $+0.72^*$ & $+0.37^\dagger$   & $+0.84^*$ & $-0.67^*$ & $+0.98^*$ & $-0.90^*$ \\
    \midrule
    \textbf{Composite} & $+0.94^*$ & $-0.14$        & $+0.97^*$ & $-0.70^*$        & $+0.55^*$ & $-0.41$ & $+0.92^*$ & $-0.86^*$ \\
    \bottomrule
  \end{tabular}
  }
  \vspace{-15pt}
\end{table}

\begin{figure}[t]
  \centering
  \vspace{-20pt}
  \includegraphics[width=\columnwidth]{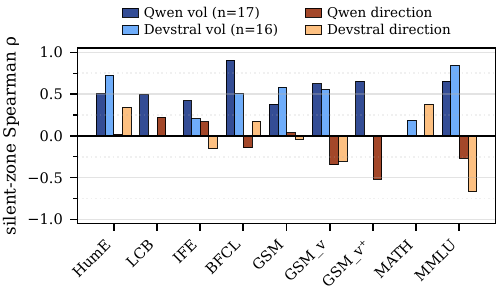}
\vspace{-25pt}
  \caption{Silent-zone Spearman $\rho(\text{KLD}, \cdot)$ per benchmark for volume (blue) and direction $f$ (orange).}
  \label{fig:vol-dir}
\vspace{-15pt}
\end{figure}

Table~\ref{tab:vol-dir} and Figure~\ref{fig:vol-dir} show the same decomposition on both cohorts. In both Qwen and Devstral, KLD consistently tracks disagreement volume: in the silent zone, $\rho(\text{KLD}, \text{volume})$ is positive on almost every benchmark, with Qwen spanning $\rho \in [+0.38,+0.91]$ and a composite $\rho=+0.94$ ($p<0.001$). On Devstral, the per-benchmark range is wider (two small-magnitude cases on IFEval and MATH-500) and the composite is weaker at $+0.55$ ($p=0.03$). Direction behaves differently. Its correlation with KLD is weak and task-dependent in the silent zone, but becomes strongly negative in the lossy zone, where large displacement is coupled with degraded quality. This explains the full-cohort KLD-quality correlation: it is driven mainly by degraded models, where KLD captures both how far the quant moved and the fact that the movement is usually harmful. Once we control for volume, the residual KLD signal on composite score does not reach significance in either cohort: partial $\rho=-0.35$ ($p=0.17$) for Qwen and $\rho=-0.20$ ($p=0.46$) for Devstral.

\vspace{-0.5em}
\paragraph{Why task matters: }
The task-dependent direction signal has a simple interpretation. Math reasoning often depends on locally determined token choices, where small deviations can compound through the solution. Code correctness is more structural: many token-level-different programs can implement the same specification, so moving away from the BF16 token distribution doesn't necessarily imply failure. On code-specialized models, where the BF16 distribution is more task-aligned, drift can become more consistently harmful.

In short, KLD measures how far the quantized model moves from the reference. Score depends on whether that movement goes in a useful direction.

\section{Limitations and Practical Implications}
\label{sec:limitation}

Our results should be read as a study of fidelity metrics for quantized-model selection, not as a claim that KLD/PPL are useless. In the lossy zone, they remain effective coarse degradation detectors. The failure mode is narrower: among near-baseline quants, where practitioners choose between plausible candidates, we do not observe a reliable ranking signal.

The silent-zone thresholds are descriptive and cohort-specific: they depend on the reference model, quantization methods, benchmark suite, and KLD setup. We therefore do not propose $\tau=0.064$ or $\tau=0.027$ as general decision rules. Our analysis is limited to two model families and a LiveCodeBench-focused routing study. Because the near-baseline zones are small ($n\approx16$--$22$), the wide confidence intervals in Appendix~\ref{sec:ci} reflect uncertainty around the silent-zone estimates, not evidence that the true correlation is exactly zero.

In practice, fidelity metrics should be treated as diagnostics rather than model-ranking criteria. High KLD/PPL can flag damaged models, but once models are near-baseline, smaller values do not necessarily imply better task quality; downstream evaluation, hardware fit, throughput, and workload-specific constraints remain essential.

\section{Conclusion}
\label{sec:conclusion}

We evaluated whether KLD and PPL can serve as proxies for downstream quality when selecting quantized LLMs. Across two cohorts, KLD appears predictive over the full quality range, but this signal is driven mainly by degraded low-bpw variants. In the near-baseline silent zone, the correlation collapses across model families and metric variants, and per-prompt KLD does not rescue model selection. The volume--direction decomposition explains why: KLD tracks how often a quantized model disagrees with the reference, but benchmark quality depends on whether those disagreements help or hurt. KLD measures displacement, not direction. For deployment, fidelity metrics should filter damaged candidates, not rank near-baseline ones.

\bibliography{custom}
\appendix

\section{Qwen MMLU Format Artifact}
\label{sec:mmlu}

We exclude Qwen MMLU from the Qwen composite quality score because of a prompt-format artifact rather than a lack of knowledge. MMLU is administered with a 5-shot prompt that instructs the model to output only the final answer choice. On this prompt, Qwen3.6 strangely does not follow the instruction: rather than emitting the answer-only format shown by the few-shot exemplars, it begins to reason in its response. Manual inspection of the generations indicates that the model generally knows the correct answer, but because its output does not match the expected answer-only format, the benchmark's answer extraction often fails to recover a valid choice and the response is scored as incorrect.

Because this behavior is present even in the BF16 reference, it lowers the absolute MMLU score of every quant in the cohort and is not a quantization effect. Including it would add a large, format-driven offset common to the whole cohort and unrelated to the fidelity question we study. We therefore report Qwen MMLU separately (Table~\ref{tab:silent-zone}) and omit it from the Qwen composite, while retaining MMLU in the Devstral composite, where this artifact does not occur.

\section{Top-\texorpdfstring{$k$}{k} and Response-Only KLD Measurement}
\label{sec:kld-topk}

This appendix specifies the two KLD variants reported as \texttt{KLD
val-response} in Table~\ref{tab:metric-invariant}: the \emph{top-$k$}
divergence and the \emph{response-only} (SFT-style) measurement protocol.

\subsection{Full-vocabulary KLD}
\label{sec:kld-full}

At a single position, an autoregressive model defines a next-token
distribution over the vocabulary $V$. Let $p \equiv p_{\mathrm{BF16}}$ be the
distribution of the BF16 reference and $q \equiv p_{\mathrm{quant}}$ that of the
candidate quant for the same context (the notation of
Section~\ref{sec:setup}). The Kullback--Leibler divergence of $q$ from $p$ is
\begin{equation}
  \mathrm{KL}(p \,\|\, q)
  \;=\; \sum_{i \in V} p(i)\,\log \frac{p(i)}{q(i)} ,
  \label{eq:kld-full}
\end{equation}
which is $0$ iff $q \equiv p$ and grows as $q$ relocates probability mass. The
expectation is taken under the reference $p$, so the metric penalizes the
quant precisely where the reference places mass. Our headline metric is the
mean of \eqref{eq:kld-full} over all scored positions.

\subsection{Top-\texorpdfstring{$k$}{k} KLD}
\label{sec:kld-topk-def}

Not all next-token predictions are equally important. In practice, when
generating from an LLM we sample from only the top few candidate tokens, so the
overwhelming majority of the vocabulary is never actually drawn. Including the
probabilities of those never-sampled tokens in \eqref{eq:kld-full} measures
agreement on a tail that has no bearing on what the model would generate. What
matters instead is whether the quant agrees with the reference on the handful
of tokens the reference would plausibly sample. We therefore restrict the
measurement to the top $k$ tokens of each distribution and renormalize.

Let $\mathcal{S}_p$ and $\mathcal{S}_q$ be the index sets of the $k$
highest-probability tokens under $p$ and $q$ respectively. Each side is
renormalized over \emph{its own} support,
\begin{equation}
  \tilde p(i) = \frac{p(i)}{Z_p},
  \qquad
  \tilde q(i) = \frac{q(i)}{Z_q},
  \label{eq:renorm}
\end{equation}
with $Z_p = \sum_{j \in \mathcal{S}_p} p(j)$ and
$Z_q = \sum_{j \in \mathcal{S}_q} q(j)$; the normalizers $\log Z_p$ and
$\log Z_q$ are computed in log-space for numerical stability. The top-$k$
divergence is summed over the \emph{reference's} support $\mathcal{S}_p$,
\begin{equation}
  \mathrm{KL}_k(p \,\|\, q)
  \;=\; \sum_{i \in \mathcal{S}_p} \tilde p(i)\,
        \bigl(\log \tilde p(i) - \log \tilde q(i)\bigr).
  \label{eq:kld-topk}
\end{equation}

The subtlety is the term $\log \tilde q(i)$ when a reference-important token
$i \in \mathcal{S}_p$ is absent from the quant's own top-$k$ set,
$i \notin \mathcal{S}_q$: the quant has effectively abandoned a token the
reference cares about. Since $\tilde q$ is renormalized over $\mathcal{S}_q$, it
assigns zero mass to any such token, so $\log \tilde q(i) = -\infty$ and the
corresponding term, and hence the entire divergence, would blow up to
$+\infty$. To keep the metric finite, we replace this $-\infty$ with a fixed
log-probability floor,
\begin{equation}
  \log \tilde q(i) =
  \begin{cases}
    \log q(i) - \log Z_q, & i \in \mathcal{S}_q, \\[2pt]
    c_{\min}, & i \notin \mathcal{S}_q,
  \end{cases}
  \label{eq:floor}
\end{equation}
with floor $c_{\min} = -10$. The floor caps the contribution of a missing token at
$\tilde p(i)\,\bigl(\log \tilde p(i) - c_{\min}\bigr)$, so a single abandoned
token incurs a large but bounded penalty rather than diverging. This design
has two properties:
\begin{itemize}[itemsep=1pt,topsep=2pt,leftmargin=1.2em]
  \item \textbf{Exact reduction.} When the two top-$k$ sets coincide,
    $\mathcal{S}_p = \mathcal{S}_q$, every term uses the renormalized
    $\log \tilde q(i)$ and \eqref{eq:kld-topk} reduces to the plain
    renormalized top-$k$ KLD.
  \item \textbf{Mass-relocation sensitivity.} When the supports diverge, the
    floored terms expose the case where the quant moved mass \emph{outside}
    the reference's top-$k$, which a renormalized-only metric would hide.
\end{itemize}

Beyond discarding irrelevant tail mass, the top-$k$ restriction has a large
practical benefit: the two-pass protocol of Section~\ref{sec:kld-sft} must
cache the reference distribution at every scored position, and storing only $k$
tokens instead of the full vocabulary shrinks this footprint by orders of
magnitude. For Qwen3.6-35B-A3B, whose vocabulary is $|V| = 248{,}320$ tokens, a
full next-token distribution stored as 16-bit quantized log-probs occupies
$\approx 485$\,KB per position; the top-$60$ record (a 32-bit index and a
32-bit log-prob per token) takes only $480$\,bytes, a $\sim\!1000\times$
reduction, rising to $\sim\!3100\times$ at $k = 20$. Over a million scored
response tokens this is the difference between roughly $500$\,GB and under half
a gigabyte of cached reference logits.

The same per-position quantity \eqref{eq:kld-topk} can be evaluated either over
a plain-text corpus (e.g.\ WikiText) or, as below, over response tokens only.
We report it at $k \in \{20, 40, 60\}$, with $k = 60$ as the default stored
support; the smaller report points reuse the prefix of the sorted top-$60$
list.

\subsection{Response-only (SFT-style) KLD}
\label{sec:kld-sft}

For instruction-tuned evaluation we care about divergence on the model's
\emph{answer}, not on the prompt it was merely conditioned on. Averaging KLD
over every position dilutes the answer-region signal with prompt tokens the
model never had to generate. The response-only protocol measures divergence
\textbf{only on response-token predictions}, while keeping the full prompt in
the conditioning context.

Each corpus sample is tokenized as a prompt followed by a response, where the
response is either an on-policy BF16 continuation (default) or a curated
reference answer. Writing the concatenated sequence as
\begin{equation}
  x \;=\; \bigl[\,\underbrace{x_1, \dots, x_m}_{\text{prompt}},\;
                  \underbrace{x_{m+1}, \dots, x_n}_{\text{response}}\,\bigr] ,
  \label{eq:concat}
\end{equation}
the sample is decoded as a single sequence, so the full prompt populates the
KV-cache. However, divergence is evaluated only at positions whose next token
is a response token: with position $t$ predicting $x_{t+1}$, this is the set
$\{m, m{+}1, \dots, n{-}1\}$. The prompt positions $\{1, \dots, m{-}1\}$
contribute context but no divergence terms.

The metric is computed in two passes over the same corpus. \emph{Pass~1}
(BF16 reference) stores, at each scored response position, the reference's
top-$k$ indices and their full-softmax log-probabilities, and optionally the
full distribution, when the full-vocabulary KLD of \eqref{eq:kld-full} is also
desired. \emph{Pass~2} (quant) runs the candidate on the identical token
sequences, recomputes its own top-$k$ at each scored position, and evaluates
$\mathrm{KL}_k(p \,\|\, q)$ from \eqref{eq:kld-topk} against the stored
reference record. Aggregating over all response tokens of all samples,
\begin{equation}
  \overline{\mathrm{KL}_k}
  \;=\; \frac{1}{N_{\text{resp}}}
        \sum_{s}\; \sum_{t=m_s}^{\,n_s - 1}
        \mathrm{KL}_k\!\bigl(p^{(s)}_t \,\|\, q^{(s)}_t\bigr),
  \label{eq:kld-agg}
\end{equation}
where $s$ indexes samples, $p^{(s)}_t, q^{(s)}_t$ are the reference and quant
next-token distributions at position $t$ of sample $s$, and
$N_{\text{resp}} = \sum_{s} (n_s - m_s)$ is the total response-token count. The
result reflects
only how the quant behaves where it must actually produce the answer,
conditioned on the true prompt: the measurement reported as \texttt{KLD
val-response} in Table~\ref{tab:metric-invariant}.

\section{A Quality-Defined Robustness Check}
\label{sec:noncircular}

In the main paper, we defined the silent zone using a descriptive cohort-specific KLD threshold $\tau$ marking the near-baseline cluster, chosen so the within-zone correlation is statistically non-significant ($p>0.10$) (Section~\ref{sec:silent}). In this appendix, we confirm that the silent-zone collapse is not an artifact of that threshold choice.

\subsection{Quality-defined silent zone}

We repeat the analysis with a KLD-threshold-free definition based only on downstream quality. Let $S_q$ denote the composite score of quantized model $q$, and let $S_{\mathrm{BF16}}$ denote the composite score of the corresponding BF16 reference. For a maximum relative quality drop $\delta$, expressed as a fraction, we define $q$ as near-baseline if
\begin{equation}
S_q \ge (1-\delta) S_{\mathrm{BF16}} .
\label{eq:nb}
\end{equation}

Our main quality-defined split uses $\delta=0.01$, corresponding to a relative quality drop of at most $1\%$. Equivalently, a quant is considered near-baseline if it retains at least $99\%$ of the corresponding BF16 reference's composite score. This criterion does not use KLD or any KLD--quality correlation. Under this definition, the near-baseline zone contains 18 Qwen quants and 22 Devstral quants; the remaining 10 and 19 quants, respectively, are assigned to the lossy zone (Table~\ref{tab:noncircular}).

\begin{table}[t]
\centering
\small
\setlength{\tabcolsep}{4pt}
\caption{Silent-zone defined purely by composite score.}
\label{tab:noncircular}
\begin{tabular}{lrr}
\toprule
\textbf{Quantity} & \textbf{Qwen} & \textbf{Devstral} \\
\midrule
Full-cohort $\rho$                 & $-0.72^*$ & $-0.86^*$ \\
\midrule
\multicolumn{3}{l}{\textit{Quality-defined silent zone ($\geq 99\%$ of BF16)}} \\
Near-baseline quants $n$           & 18        & 22 \\
KLD range covered                  & $\leq 0.094$ & $\leq 0.060$ \\
Silent-zone $\rho$                 & $-0.07$   & $-0.23$ \\
Silent-zone $p$                    & $0.80$    & $0.31$ \\
\bottomrule
\end{tabular}
\end{table}

\subsection{The collapse persists}

Within this quality-defined zone, the KLD--quality correlation remains non-significant: $\rho=-0.07$ ($p=0.80$) on Qwen and $\rho=-0.23$ ($p=0.31$) on Devstral, compared with full-cohort correlations of $-0.72$ and $-0.86$ (both $p<0.001$). The collapse therefore does not depend on the KLD-based threshold used in the main text. On Qwen the quality-defined zone is also wider in KLD than the main silent zone, reaching $KLD \leq 0.094$ on Qwen and $KLD \leq 0.060$ on Devstral, compared to $\tau=0.064$ and $\tau=0.027$.

\subsection{Sensitivity to the quality bar}

Table~\ref{tab:noncircular-sens} varies the $\delta$ quality bar (relative drop from BF16) from $0.5\%$ to $3.0\%$. The KLD--quality correlation remains non-significant on both cohorts through a $1.5\%$ bar. As the bar is relaxed further, degraded quants enter the near-baseline set and the expected negative correlation re-emerges, reaching significance at $\delta = 2.0\%$ on Devstral ($\rho=-0.41$, $p=0.04$) and $\delta = 2.5\%$ on Qwen ($\rho=-0.55$, $p<0.01$). This is consistent with the main result: KLD becomes informative primarily when sufficiently degraded models enter the comparison set, but not among near-baseline candidates.

\subsection{Metric-invariance is preserved}

Repeating the quality-defined check across the measurement variants in Table~\ref{tab:metric-invariant} gives the same qualitative result (Table~\ref{tab:metric-invariant-quality}). KLD aggregations, PPL formulations, top-1 agreement, long-context variants, and response-only variants all remain non-significant in the quality-defined zone (Qwen $|\rho|\leq 0.13$; Devstral $|\rho|\leq 0.25$), confirming that the conclusion is not specific to either the metric definition or the KLD threshold.

\begin{table}[t]
\centering
\small
\setlength{\tabcolsep}{4pt}
\caption{Sensitivity of the quality-defined silent zone to the deployment bar $\delta$ (maximum relative drop from BF16). $n$ is the number of near-baseline quants. The collapse is non-significant on both cohorts through $\delta=1.5\%$; the $\delta=1\%$ operating point is in bold.}
\label{tab:noncircular-sens}
\begin{tabular}{rrrr|rrr}
\toprule
& \multicolumn{3}{c|}{\textbf{Qwen}} & \multicolumn{3}{c}{\textbf{Devstral}} \\
\cmidrule(lr){2-4}\cmidrule(lr){5-7}
\textbf{$\delta\%$} & \textbf{$n$} & \textbf{$\rho$} & \textbf{$p$} & \textbf{$n$} & \textbf{$\rho$} & \textbf{$p$} \\
\midrule
0.5 & 14 & $+0.11$ & 0.71 & 16 & $+0.02$ & 0.93 \\
\textbf{1.0} & \textbf{18} & $\mathbf{-0.07}$ & \textbf{0.80} & \textbf{22} & $\mathbf{-0.23}$ & \textbf{0.31} \\
1.5 & 18 & $-0.07$ & 0.80 & 23 & $-0.30$ & 0.16 \\
2.0 & 20 & $-0.25$ & 0.29 & 25 & $-0.41$ & 0.04 \\
2.5 & 24 & $-0.55$ & 0.00 & 26 & $-0.47$ & 0.02 \\
3.0 & 24 & $-0.55$ & 0.00 & 26 & $-0.47$ & 0.02 \\
\bottomrule
\end{tabular}
\end{table}

\begin{table}[t]
  \centering
  \footnotesize
  \setlength{\tabcolsep}{1pt}
  \caption{Metric-invariance under the quality-defined silent zone (composite $\geq 99\%$ of BF16).}
  \label{tab:metric-invariant-quality}
  \resizebox{\linewidth}{!}{
  \begin{tabular}{lrr|rr}
    \toprule
    & \multicolumn{2}{c}{\textbf{Qwen3.6-35B-A3B}} & \multicolumn{2}{c}{\textbf{Devstral-Small-2-24B}} \\
    \cmidrule(lr){2-3} \cmidrule(lr){4-5}
    \textbf{Metric variant} & \textbf{full} & \textbf{$\geq$99\%} & \textbf{full} & \textbf{$\geq$99\%} \\
    \midrule
    KLD mean wiki c512        & $-0.72^*$ & $-0.07$ & $-0.86^*$ & $-0.23$ \\
    KLD median wiki c512      & $-0.69^*$ & $-0.04$ & $-0.86^*$ & $-0.23$ \\
    KLD p90 wiki c512         & $-0.71^*$ & $-0.05$ & $-0.86^*$ & $-0.23$ \\
    KLD p99 (tail) wiki c512  & $-0.75^*$ & $-0.13$ & $-0.86^*$ & $-0.21$ \\
    mean $\Delta\log$-PPL wiki c512 & $-0.68^*$ & $-0.05$ & $-0.87^*$ & $-0.25$ \\
    PPL ratio (q/base) wiki c512 & $-0.68^*$ & $-0.05$ & $-0.87^*$ & $-0.25$ \\
    PPL diff wiki c512        & $-0.68^*$ & $-0.05$ & $-0.87^*$ & $-0.25$ \\
    top-1 agreement \% wiki c512 & $+0.70^*$ & $+0.02$ & $+0.86^*$ & $+0.24$ \\
    KLD wiki c8k (long context) & $-0.72^*$ & $-0.07$ & $-0.86^*$ & $-0.22$ \\
    PPL ratio wiki c8k        & $-0.69^*$ & $-0.04$ & $-0.86^*$ & $-0.20$ \\
    top-1 agreement \% wiki c8k & $+0.69^*$ & $+0.03$ & $+0.86^*$ & $+0.24$ \\
    KLD val-wiki c512         & $-0.71^*$ & $-0.04$ & ---       & ---     \\
    KLD val-response top-$k$=60 & $-0.73^*$ & $-0.06$ & $-0.86^*$ & $-0.22$ \\
    KLD val-response top-$k$=20 & $-0.73^*$ & $-0.06$ & $-0.87^*$ & $-0.23$ \\
    \bottomrule
  \end{tabular}
  }
\end{table}

\subsection{A check, not a selection method}

The quality-defined zone is not intended as an operational selection rule: practitioners do not know downstream quality before running benchmarks. It is used only as a robustness check. The main text therefore retains the KLD-based $\tau$ as a descriptive KLD-space characterization of the collapse, while this appendix verifies that the collapse is also present under a quality-only definition.

\section{Internal Validation Dataset}
\label{sec:dataset}

The \texttt{KLD val-response} rows in Table~\ref{tab:metric-invariant} are computed on held-out validation slices from internal instruction mixtures. These measurements are used only as robustness checks: the headline KLD metric is computed on WikiText with 512-token context. The purpose of the validation slices is to test whether the silent-zone collapse persists under a prompt distribution closer to instruction-following use.

For each reference model, we build a separate validation slice by stratified sampling from source datasets covering instruction following, multi-turn chat, code generation, math reasoning, tool calling, and multilingual prompts. The slices are not used for training any quantized model in this study. They are used only to compute response-only KLD against the corresponding BF16 reference.

\begin{table}[t]
\centering
\small
\setlength{\tabcolsep}{4pt}
\caption{Internal validation slices used for the \texttt{KLD val-response} robustness checks in Table~\ref{tab:metric-invariant}.}
\label{tab:sft-val-composition}
\vspace{-5pt}
\resizebox{\linewidth}{!}{
\begin{tabular}{lrr}
\toprule
\textbf{Property} & \textbf{Qwen} & \textbf{Devstral} \\
\midrule
Samples (scored)          & 1{,}662 & 1{,}364 \\
Median prompt length      & 174     & 168 \\
Tool-bearing samples (\%) & 9.3     & 20.8 \\
Thinking-mode samples (\%) & 24.8    & --- \\
\bottomrule
\end{tabular}
}
\end{table}
\begin{table}[t]
\centering
\small
\setlength{\tabcolsep}{4pt}
\caption{Bootstrap $95\%$ confidence intervals for headline Spearman correlations. Intervals are percentile bootstrap intervals over quantized models.}
\vspace{-5pt}
\label{tab:ci-corr}
\resizebox{\linewidth}{!}{
\begin{tabular}{llrrr}
\toprule
\textbf{Analysis} & \textbf{Cohort} & \textbf{$n$} & \textbf{$\rho$} & \textbf{$95\%$ CI} \\
\midrule
\multirow{2}{*}{Full cohort} & Qwen     & 28 & $-0.72$ & $[-0.89,-0.39]$ \\
& Devstral & 41 & $-0.86$ & $[-0.95,-0.68]$ \\
\midrule
KLD-defined & Qwen     & 17 & $+0.00$ & $[-0.53,+0.54]$ \\
silent zone& Devstral & 16 & $-0.24$ & $[-0.63,+0.33]$ \\
\midrule
\multirow{2}{*}{Lossy zone} & Qwen     & 11 & $-0.80$ & $[-1.00,-0.29]$ \\
& Devstral & 25 & $-0.94$ & $[-0.99,-0.81]$ \\
\midrule
Quality-defined  & Qwen     & 18 & $-0.07$ & $[-0.55,+0.47]$ \\
silent zone & Devstral & 22 & $-0.23$ & $[-0.63,+0.29]$ \\
\midrule
Volume& Qwen     & 17 & $+0.94$ & $[+0.75,+1.00]$ \\
silent zone& Devstral & 16 & $+0.55$ & $[+0.01,+0.88]$ \\
\midrule
Partial $\rho$ controlling& Qwen     & 17 & $-0.35$ & $[-0.75,+0.32]$ \\
for volume & Devstral & 16 & $-0.20$ & $[-0.61,+0.47]$ \\
\bottomrule
\end{tabular}
}
\vspace{-10pt}
\end{table}

\section{Confidence Intervals for Headline Correlations}
\label{sec:ci}

To further estimate uncertainty for the headline Spearman correlations, we perform a nonparametric bootstrap over the quantized models. For each cohort and split, each bootstrap replicate samples $n$ models with replacement from the corresponding set and recomputes $\rho(\mathrm{KLD}, \mathrm{composite})$. We used $B=10{,}000$ bootstrap replicates with a fixed seed and report percentile $95\%$ confidence intervals. We computed the intervals on the same data and zone definitions as the corresponding point estimates.

Because the silent-zone subsets are small, these intervals should be interpreted as uncertainty estimates rather than evidence that the true correlation is exactly zero. The purpose of this analysis is to test whether the strong full-cohort KLD--quality relationship is also supported inside the near-baseline region.

\begin{table}[t]
\centering
\footnotesize
\setlength{\tabcolsep}{4pt}
\caption{Qwen3.6-35B-A3B cohort ($n=28$, $\tau=0.064$).}
\vspace{-5pt}
\label{tab:models-qwen}
\resizebox{\linewidth}{!}{
\begin{tabular}{llrrrl}
\toprule
\textbf{Publisher} & \textbf{Model} & \textbf{bpw} & \textbf{KLD} & \textbf{comp} & \textbf{zone} \\
\midrule
unsloth & UD-Q5\_K\_XL & 6.14 & 0.0082 & 0.724 & silent \\
AesSedai & Q5\_K\_M & 6.06 & 0.0083 & 0.718 & silent \\
mudler & APEX-Balanced & 5.91 & 0.0106 & 0.717 & silent \\
bartowski & Q5\_K\_L & 5.84 & 0.0110 & 0.720 & silent \\
unsloth & UD-Q4\_K\_XL & 5.16 & 0.0135 & 0.728 & silent \\
AesSedai & Q4\_K\_M & 5.11 & 0.0140 & 0.719 & silent \\
mudler & APEX-Quality & 5.26 & 0.0140 & 0.727 & silent \\
bartowski & Q4\_K\_L & 5.02 & 0.0196 & 0.717 & silent \\
bartowski & Q4\_K\_M & 4.93 & 0.0208 & 0.722 & silent \\
bartowski & IQ4\_XS & 4.34 & 0.0285 & 0.725 & silent \\
unsloth & UD-IQ4\_NL & 4.16 & 0.0334 & 0.726 & silent \\
AesSedai & IQ4\_XS & 4.06 & 0.0335 & 0.717 & silent \\
unsloth & UD-IQ4\_XS & 4.09 & 0.0336 & 0.724 & silent \\
unsloth & UD-Q3\_K\_XL & 3.89 & 0.0434 & 0.715 & silent \\
mudler & APEX-Compact & 3.99 & 0.0451 & 0.707 & silent \\
ByteShape & IQ4\_XS-4.15bpw & 4.15 & 0.0504 & 0.731 & silent \\
ByteShape & Q4\_K\_S-4.22bpw & 4.22 & 0.0582 & 0.723 & silent \\
ByteShape & IQ4\_XS-3.93bpw & 3.93 & 0.0689 & 0.721 & lossy \\
unsloth & UD-IQ3\_S & 3.15 & 0.0842 & 0.708 & lossy \\
AesSedai & IQ3\_S & 3.13 & 0.0850 & 0.705 & lossy \\
ByteShape & Q4\_K\_S-3.80bpw & 3.80 & 0.0944 & 0.714 & lossy \\
ByteShape & Q3\_K\_S-3.39bpw & 3.39 & 0.0985 & 0.705 & lossy \\
mudler & APEX-Mini & 3.30 & 0.0994 & 0.695 & lossy \\
ByteShape & IQ3\_S-3.48bpw & 3.48 & 0.1223 & 0.706 & lossy \\
ByteShape & IQ3\_S-3.00bpw & 3.00 & 0.1273 & 0.706 & lossy \\
ByteShape & Q3\_K\_S-2.71bpw & 2.71 & 0.1917 & 0.685 & lossy \\
ByteShape & Q3\_K\_S-2.69bpw & 2.69 & 0.2092 & 0.674 & lossy \\
ByteShape & IQ2\_S-2.17bpw & 2.17 & 0.2733 & 0.654 & lossy \\
\bottomrule
\end{tabular}
}
\vspace{-15pt}
\end{table}

Table~\ref{tab:ci-corr} supports the same interpretation as our main results. The full-cohort and lossy-zone KLD--quality intervals exclude zero on both cohorts. In contrast, the KLD--quality intervals for both the KLD-defined silent zone and the quality-defined silent zone are wide and straddle zero. Thus, the silent-zone result should be read as an absence of detectable KLD--quality ranking signal in the near-baseline region, not as a precise estimate that the true correlation is exactly zero. The volume correlations remain positive, especially on Qwen, while the partial correlations controlling for volume also straddle zero. This is consistent with the volume--direction account: KLD reliably tracks displacement volume, but its residual ranking signal for downstream quality is not detectable among near-baseline candidates.

\section{Full Model List}
\label{sec:model-list}

We list every quantized model in the two cohorts in
Tables~\ref{tab:models-qwen} and~\ref{tab:models-devstral}, sorted by KLD. All included quants are official public GGUF releases from their respective publishers.

\begin{table}[t]
\centering
\footnotesize
\setlength{\tabcolsep}{4pt}
\caption{Devstral-Small-2-24B cohort ($n=41$, $\tau=0.027$).}
\label{tab:models-devstral}
\begin{tabular}{llrrrl}
\toprule
\textbf{Publisher} & \textbf{Model} & \textbf{bpw} & \textbf{KLD} & \textbf{comp} & \textbf{zone} \\
\midrule
unsloth & Q8\_0 & 8.50 & 0.0008 & 0.696 & silent \\
unsloth & UD-Q8\_K\_XL & 9.84 & 0.0009 & 0.700 & silent \\
unsloth & UD-Q6\_K\_XL & 7.05 & 0.0019 & 0.694 & silent \\
bartowski & Q6\_K\_L & 6.67 & 0.0025 & 0.695 & silent \\
unsloth & Q6\_K & 6.56 & 0.0027 & 0.711 & silent \\
bartowski & Q6\_K & 6.56 & 0.0027 & 0.697 & silent \\
bartowski & Q5\_K\_L & 5.83 & 0.0069 & 0.699 & silent \\
unsloth & Q5\_K\_M & 5.69 & 0.0070 & 0.690 & silent \\
bartowski & Q5\_K\_M & 5.69 & 0.0071 & 0.692 & silent \\
unsloth & UD-Q5\_K\_XL & 5.69 & 0.0072 & 0.692 & silent \\
unsloth & Q5\_K\_S & 5.53 & 0.0084 & 0.695 & silent \\
bartowski & Q5\_K\_S & 5.53 & 0.0086 & 0.695 & silent \\
unsloth & UD-Q4\_K\_XL & 4.92 & 0.0200 & 0.695 & silent \\
bartowski & Q4\_K\_M & 4.86 & 0.0219 & 0.694 & silent \\
unsloth & Q4\_K\_M & 4.86 & 0.0222 & 0.696 & silent \\
unsloth & Q4\_1 & 5.04 & 0.0266 & 0.697 & silent \\
unsloth & Q4\_K\_S & 4.60 & 0.0274 & 0.691 & lossy \\
bartowski & IQ4\_NL & 4.57 & 0.0278 & 0.685 & lossy \\
unsloth & IQ4\_NL & 4.57 & 0.0279 & 0.696 & lossy \\
bartowski & IQ4\_XS & 4.33 & 0.0284 & 0.691 & lossy \\
unsloth & IQ4\_XS & 4.33 & 0.0286 & 0.697 & lossy \\
unsloth & Q4\_0 & 4.58 & 0.0387 & 0.690 & lossy \\
ByteShape & IQ4\_XS-4.04bpw & 4.04 & 0.0473 & 0.698 & lossy \\
unsloth & UD-Q3\_K\_XL & 4.02 & 0.0600 & 0.691 & lossy \\
unsloth & Q3\_K\_M & 3.89 & 0.0684 & 0.683 & lossy \\
ByteShape & IQ3\_S-3.47bpw & 3.47 & 0.1181 & 0.685 & lossy \\
unsloth & Q3\_K\_S & 3.53 & 0.1260 & 0.663 & lossy \\
ByteShape & IQ3\_S-3.19bpw & 3.19 & 0.2062 & 0.672 & lossy \\
unsloth & UD-Q2\_K\_XL & 3.15 & 0.2579 & 0.636 & lossy \\
unsloth & UD-IQ3\_XXS & 3.19 & 0.2592 & 0.655 & lossy \\
ByteShape & IQ3\_S-2.96bpw & 2.96 & 0.2733 & 0.658 & lossy \\
unsloth & Q2\_K\_L & 3.07 & 0.2913 & 0.635 & lossy \\
unsloth & Q2\_K & 3.01 & 0.2936 & 0.631 & lossy \\
ByteShape & IQ3\_S-2.78bpw & 2.78 & 0.3617 & 0.639 & lossy \\
ByteShape & IQ3\_S-2.67bpw & 2.67 & 0.4425 & 0.620 & lossy \\
unsloth & UD-IQ2\_M & 2.79 & 0.5013 & 0.613 & lossy \\
ByteShape & IQ2\_S-2.43bpw & 2.43 & 0.6525 & 0.572 & lossy \\
ByteShape & IQ2\_S-2.34bpw & 2.34 & 0.7991 & 0.536 & lossy \\
unsloth & UD-IQ2\_XXS & 2.29 & 1.2795 & 0.348 & lossy \\
unsloth & UD-IQ1\_M & 2.04 & 2.7823 & 0.128 & lossy \\
unsloth & UD-IQ1\_S & 1.88 & 3.5199 & 0.089 & lossy \\
\bottomrule
\end{tabular}
\end{table}

\end{document}